\documentclass[cameraready]{Interspeech}
\title{COALA: Robust Contextualized Speech-augmented Language Modeling for ASR via Contrastive Regularizer and Biasing Score Estimation}

\author[affiliation={}]{Jhih-Rong}{Guo}
\author[affiliation={}]{Bi-Cheng}{Yan}
\author[affiliation={}]{Tien-Hong}{Lo}
\author[affiliation={}]{Berlin}{Chen}

\address{
    National Taiwan Normal University, Taiwan
}

\email{\{jhihrong, 80847001s, teinhonglo, berlin\}@ntnu.edu.tw}

\keywords{speech recognition, contextual biasing, biasing scoring, contrastive learning}

\usepackage{comment}

\usepackage{stfloats} 
\usepackage{multirow} 
\usepackage{makecell} 
\usepackage{threeparttable} 

\begin{document}

\maketitle

\begin{abstract}
    Contextual biasing seeks to integrate external knowledge into automatic speech recognition (ASR) systems to accurately recognize domain-specific entities. In this paper, we propose COALA\footnote{The source code and experimental setup for this study are available at https://github.com/Guo0911/COALA.} (\textbf{CO}ntextualized \textbf{A}SR \textbf{L}everaging bi\textbf{A}sing scoring), a robust framework designed to enhance speech-augmented language models (SLMs) in complex multi-entity scenarios. Considering the inherent context-window limitations of SLMs, identifying relevant target entities from a large-scale biasing list is crucial for effective recognition. To this end, COALA maps SLM latent representations into a specialized discriminative space to quantify the matching intensity between audio segments and candidate entities. Furthermore, we address the training collapse in prior study when handling multi-target utterances—where multiple rare words co-occur. Experimental results on the LibriSpeech benchmark demonstrate that COALA consistently achieves superior contextual biasing performance across various biasing list scales.
\end{abstract}

\section{Introduction}
Driven by the monolithic nature and streamlined training process, end-to-end (E2E) automatic speech recognition (ASR) systems \cite{graves2006connectionist, graves2012sequence, bahdanau2015neural, tang2023hybrid} have gained widespread attention across both academia and industry.
More recently, as witnessed by the remarkable success of large language models (LLMs) in the natural language processing community, efforts in ASR have pivoted toward leveraging LLMs for speech recognition and audio reasoning tasks.
Through the synergy of a speech encoder and a language model, speech-augmented language models (SLMs) serve as a promising paradigm for ASR \cite{chu2023qwen, tang2023salmonn, chu2024qwen2, lam2025prepending, hu2024wavllm}.

However, SLMs still struggle with uncommon, domain-specific entities such as contact names, proper nouns, and other named entities.
Accordingly, contextual biasing techniques have been widely investigated to incorporate external knowledge into ASR systems, steering ASR hypotheses toward the accurate recognition of target entities and exerting a profound impact on modern speech applications..
Existing literature on contextual biasing can be broadly categorized into two active strands of research, namely, inference-time and training-time biasing, based on the stage of external knowledge integration.
Inference-time biasing approaches, such as shallow fusion \cite{mohri2002weighted, zhao2019shallow} and on-the-fly rescoring \cite{williams2018contextual, huang2025neuralmodelcontextualbiasing}, typically construct n-gram finite state transducers (FSTs) from a curated knowledge dataset, which are then utilized to selectively bias the search space and dynamically boost the emission probability of target entities during the decoding stage, often triggered by specific prefixes (e.g., 'call' or 'play').
On a separate front, training-time biasing methods, such as attention-based biasing adapters \cite{sathyendra2022contextual, jayanthi2023retrieve} and trie-based pointer generators \cite{sun2022minimising, sun2024graph}, which are trained to align acoustic features with contextual information via location-aware attention or neural shortcuts, effectively internalizing the recognition ability to prioritize target entities from custom biasing lists in an E2E manner.

\begin{figure}[t]
  \centering
  \includegraphics[width=\linewidth]{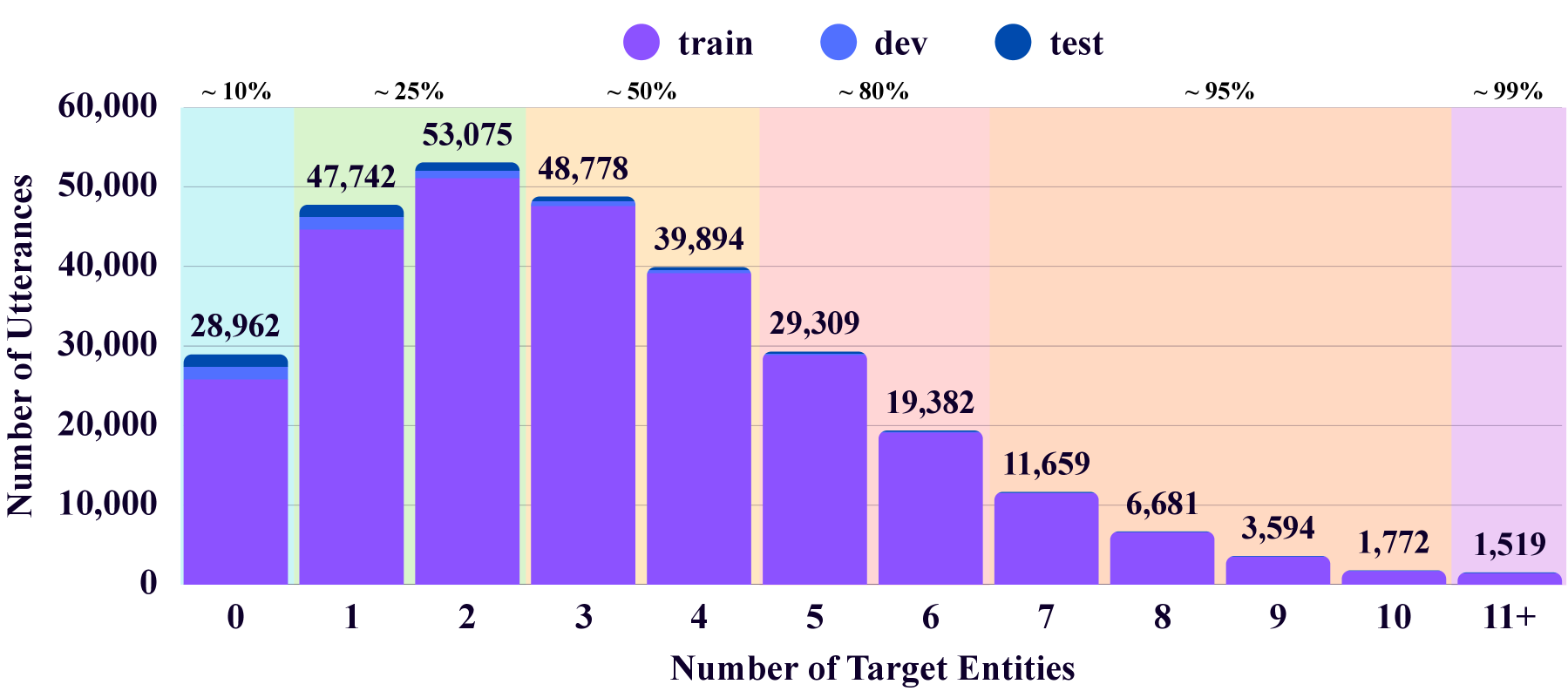}
  \caption{Distribution of utterances by number of target entities on the LibriSpeech corpus.}
  \label{fig:data_distribution}
\end{figure}

Despite the progress in inference-time and training-time biasing techniques, seamlessly integrating these methods into the evolving architectures of SLMs remains a formidable challenge.
As the scale of the biasing list expands, SLMs often suffer from substantial performance degradation due to inherent context-window limitations and the interference of excessive distractor entities.
Inspired by recent advancements in training discriminative scorers for entity filtering, we propose COALA (\textbf{CO}ntextualized \textbf{A}SR \textbf{L}everaging bi\textbf{A}sing scoring), a novel contextual biasing framework specifically tailored for SLMs.
COALA quantifies the matching intensity between acoustic feature and target entities by mapping the latent representations of SLMs into a specialized discriminative space.
This approach effectively leverages the potent capabilities of SLMs while remaining independent of the standard language model vocabulary distribution.
To further enhance the robustness of our framework, we introduce two refined objective functions: Multi-Positive Discriminative Loss (MPD-Loss) and Decoupled Point-wise Discriminative Loss (DPD-Loss).
These functions address a critical limitation of discriminative loss, which typically struggle with multi-target training data—where multiple rare words co-occur in a single utterance—or necessitate an auxiliary log loss for convergence.
By resolving gradient conflicts and enabling stable training on multi-entity data, our proposed losses effectively utilize the significant portion of complex data present in the used dataset (as illustrated in Figure \ref{fig:data_distribution}).

\section{Methods}

\begin{figure*}[htbp]
  \centering
  \includegraphics[width=\linewidth]{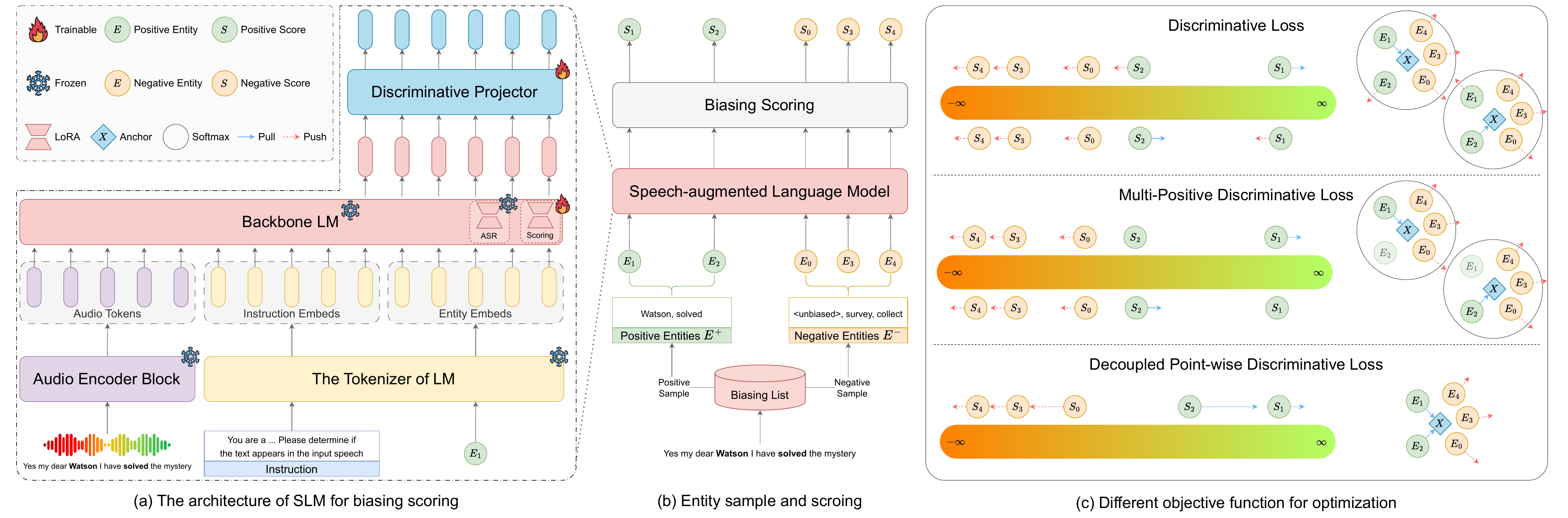}
  \caption{
      (a) The proposed COALA framework for biasing scoring, featuring a frozen backbone with trainable LoRA modules and a discriminative projector.
      (b) The sampling and scoring process where positive ($E^+$) and negative ($E^-$) entities are assigned matching intensities scores $S$.
      (c) Comparison of objective functions: discriminative loss exhibits inter-positive competition, while multi-positive discriminative loss (MPD-Loss) and decoupled point-wise discriminative loss (DPD-Loss) systematically resolve gradient conflicts through local and point-wise decoupling, respectively.
  }
  \label{fig:algorithm}
\end{figure*}

\subsection{Biasing Scoring}
As illustrated in Figure \ref{fig:algorithm} (a), we define the acoustic features extracted from the input audio via the audio encoder block as $X$, and the corresponding ground-truth transcription as $Y$.
For each utterance, a biasing list $E=\{e_0, e_1, e_2, \dots, e_M\}$ is constructed, containing $M+1$ entities.
Within this list, $e_0$ is a special token, denoted as $<$unbiased$>$, which indicates the absence of any entities in the utterance.
The biasing list $E$ comprises a set of positive entities $E^+ \subseteq Y$ and negative entities $E^- \not\subseteq Y$.
Specifically, if no entities from the list appear in the ground-truth $Y$, the $<$unbiased$>$ token serves as the single positive entity ($E^+=\{e_0\}$); otherwise, it is treated as a negative entity.
Each candidate entity $e_i$ is represented as a token sequence $e_i=\{e_{i1}, e_{i2}, \dots, e_{iL_i}\}$, where $L_i$ denotes the sequence length.
To determine the matching degree, the entity embeddings are fed into the backbone LM alongside the audio tokens and instruction embeddings.
We compute a sequence-level discriminative score $s_i$ by performing length-normalization on the token-level matching intensities, which ensures a fair comparison between entities of varying lengths:
\begin{align}
  s_i=\frac{1}{L_i} \sum_{j=1}^{L_i} f(h_{ij}) \in \mathbb{R},
  \label{equation:score}
\end{align}
where $h_{ij}$ is the last hidden state extracted from the backbone LM corresponding to the $j$-th token of entity $e_i$.
The scoring function $f(\cdot)$ is implemented as a discriminative projector, which is a multi-layer perceptron (MLP) designed to map latent representations into a specialized discriminative space:
\begin{align}
  f(h_{ij})=W_2 \cdot \text{ReLU}(W_1 \cdot h_{ij} + b_1) + b_2,
\end{align}
where $\text{ReLU}(\cdot)$ is the activation function.
$W_1 \in \mathbb{R}^{D' \times D}$ and $b_1 \in \mathbb{R}^{D'}$ denote the weight and bias of the first linear layer, while $W_2 \in \mathbb{R}^{1 \times D'}$ and $b_2 \in \mathbb{R}$ represent those of the second layer.
This MLP-based scoring mechanism allows the model to quantifies the matching intensity between acoustic feature and target entities, independent of the standard language model vocabulary distribution.

\subsection{Optimization Objective Function}
In the prior study \cite{huang2025neuralmodelcontextualbiasing}, a neural network was trained using a discriminative loss designed to ensure that the matching intensities of positive and negative entities are clearly distinguishable.
To maintain numerical stability, a softmax function is applied to normalize the scores $S=\{s_0, s_1, s_2, \dots, s_M\}$.
The discriminative loss is formally defined as:
\begin{align}
  \mathcal{L}_{disc}=\frac{-1}{|E^+|}\sum_{i \in E^+}\log{
    \frac{exp(s_i)}{\sum_{j \in E}exp(s_j)}
  }.
\end{align}
As illustrated in Figure \ref{fig:algorithm} (c), although discriminative loss effectively separates entities in single-target tasks, its global normalization imposes an unintended competition among positive entities.
Specifically, when the biasing list contains multiple positive entities, increasing the score of one positive entity $s_i$ forcedly suppresses the normalized values of others in $E^+$, leading to gradient conflicts and sub-optimal convergence.
To address this, we propose the multi-positive discriminative loss (MPD-Loss), which independently contrasts each positive entity against the set of negative entities $E^-$:
\begin{align}
  \mathcal{L}_{MPD}=\frac{-1}{|E^+|}\sum_{i \in E^+}\log{
    \frac{exp(s_i)}{exp(s_i) + \sum_{j \in E^-}exp(s_j)}
  }.
\end{align}
By localizing the softmax normalization to a positive entity and the set of negative entities, MPD-Loss eliminates the mutual exclusivity among positive entities, allowing the model to simultaneously pull multiple positive entities toward the acoustic anchor $X$ without interference.
However, MPD-Loss mitigates the competition among positive entities, it remains a relative ranking objective where the optimization of positive entities is intrinsically coupled with the negative distribution.
Under this formulation, the model focuses on ensuring the scores of positive entities are sufficiently higher than negative entities rather than calibrating its absolute magnitude.
To further decouple these interactions and provide a more robust discriminative signal, we propose the decoupled point-wise discriminative loss (DPD-Loss) that defined as:
\begin{align}
  &\mathcal{L}_{DPD}=\mathcal{P}(E^+)+\mathcal{N}(E^-),
\end{align}
where $\mathcal{P}(\cdot)$ and $\mathcal{N}(\cdot)$ are leveraging the sigmoid function $\sigma(\cdot)$ to optimize absolute matching intensities for input biasing list:
\begin{align}
  &\mathcal{P}(E)=\frac{-1}{|E|}\sum_{i \in E}\log{(\sigma(s_i))}, \\
  &\mathcal{N}(E)=\frac{-1}{|E|}\sum_{j \in E}\log{(1-\sigma(s_j))}.
\end{align}
As illustrated in Figure \ref{fig:algorithm} (c), DPD-Loss treats each entity as an independent binary classification task.
By adopting this point-wise approach, DPD-Loss eliminates the residual gradient coupling between positive and negative gradients inherent in softmax-based objectives.
This transition from relative ranking to absolute discrimination allows the model to learn a more stable and well-defined decision boundary.

\section{Experiments}
\subsection{Datasets}
We evaluate our proposed method on the LibriSpeech corpus \cite{librispeech2015}, following the contextual biasing experimental setup established in prior study \cite{le2021contextualized}.
In this setup, 5K high-frequency terms are identified as common words, while the remaining 209.2K lower-frequency terms are categorized as rare words.
For each utterance, we construct a biasing list of size $N=\{500, 1000, 5000\}$.
Each list consists of all rare words present in the utterance as positive entities, supplemented by randomly sampled rare words not appearing in the utterance as negative entities.
The models are trained on the full 960-hour training set, with dev-clean serving as the validation set for best model selection.

\subsection{Evaluation Metrics}
To assess the performance of both ASR and biasing scoring, we utilize word error rate (WER) and recall as our primary metrics.
Specifically, WER is decomposed into unbiased WER (U-WER) and biased WER (B-WER), which represent the error rates for common words and biasing words, respectively.
For the scoring task, we report two variants of recall: Recall@X and Recall\#X.
Recall@X denotes the average number of highest-scoring entities required to achieve an $X\%$ recall per utterance.
Conversely, Recall\#X represents the recall rate achieved when selecting the top-$X$ highest-scoring entities for each utterance.

\subsection{Baselines}
Beyond comparing with discriminative loss, we evaluate our approach against two representative prompt-based baselines: CTC-Filter \cite{yang2024ctc} and knowledge prompt (K-Prompt) \cite{zhang2023knowledgepromptforwhisper}.
The former utilizes coarse CTC decoding results to filter and incorporate pertinent entities into the SLM prompt.
The latter adopts a multi-pass strategy, performing fuzzy matching on initial ASR transcriptions against a knowledge base to generate entity-specific prompts for a second-pass decoding.

\subsection{Implementation Details}
Our speech-augmented language model (SLM) framework dubbed COALA integrates a pre-trained Whisper-large-v2 encoder \cite{whisper2023robust} (640M parameters) and HuggingFaceTB/SmolLM2-135M-Instruct\footnote{https://huggingface.co/HuggingFaceTB/SmolLM2-135M-Instruct} as the backbone LM, totaling 777M parameters to evaluate COALA in lightweight settings.
To enhance acoustic grounding, a dedicated CTC module is incorporated following the audio adapter.
This module generates frame-level alignments that are projected into the backbone LM embedding space as audio tokens to serve as the prefix for the instruction-tuned decoder.
Internal experiments indicate that this integration significantly improves transcription robustness by providing more precise acoustic-semantic alignment.

Experiments are conducted on a 24GB RTX 3090 GPU. The model training proceeds in two stages:
\begin{itemize}
    \item
    Stage 1 (ASR Training): We jointly train the audio adapter and CTC module from scratch while fine-tuning the backbone LM via LoRA for 10 epochs with a batch size of 4.
    To prevent over-reliance on biasing information, prompts are constructed by sampling up to 5 positive entities within a 10-entity set.
    Optimization is performed using a combination of CTC and cross-entropy losses, with weights set to 0.3 and 1.0, respectively.
    \item
    Stage 2 (Scoring Training): All previously trained parameters are frozen. A newly initialized discriminative projector and an additional LoRA module within the backbone LM are trained for 7 epochs with a batch size of 1.
    This stage utilizes a candidate list of $M=120$ entities.
    Notably, while the baseline discriminative loss requires an auxiliary log loss (weight 0.1) to converge (consistent with prior studies \cite{huang2025neuralmodelcontextualbiasing}), our proposed MPD-Loss and DPD-Loss function effectively as standalone objectives.
\end{itemize}
During inference, candidate entities are ranked by discriminative scores calculated via Equation \ref{equation:score}.
We adopt a top-$K$ strategy ($K=10$) to select the target entities for the ASR prompt, which is called biasing target identification.
To further mitigate performance degradation from negative entities, we use the score of the special token $<$unbiased$>$ as a threshold to filter low-score entities, as 99\% of our utterances contain fewer than 10 entities.

\section{Results}
\subsection{The Performance of Biasing Scoring}

\begin{table}[t]
    \caption{Performance of biasing scoring on test-clean and test-other with $N=5000$. The Bias-Loss is a joint objective that combination of the discriminative loss and log loss.}
    \label{tab:MainRetrieval}
    \centering
    \resizebox{\linewidth}{!}{
      \begin{tabular}{lccccccc}

        \toprule
        \multicolumn{1}{c}{\multirow{2}{*}{\textbf{Methods}}} &
        \multicolumn{2}{c|}{\textbf{Recall@95 ($\downarrow$)}} &
        \multicolumn{2}{c|}{\textbf{Recall\#20 (\%) ($\uparrow$)}} &
        \multicolumn{2}{c}{\textbf{Recall\#50 (\%) ($\uparrow$)}} \\
        \cmidrule{2-7}
          &
        \textbf{clean} & \multicolumn{1}{c|}{\textbf{other}} &
        \textbf{clean} & \multicolumn{1}{c|}{\textbf{other}} &
        \textbf{clean} & \textbf{other} \\
        \midrule

        CTC-Filter \cite{yang2024ctc} &
        94.3 & \multicolumn{1}{c|}{1459.0} &
        93.26 & \multicolumn{1}{c|}{83.83} &
        94.45 & 85.49 \\

        K-Prompt \cite{zhang2023knowledgepromptforwhisper} &
        593.3 & \multicolumn{1}{c|}{1532.3} &
        86.30 & \multicolumn{1}{c|}{73.88} &
        88.92 & 79.05 \\

        \midrule
        \multicolumn{2}{l}{COALA} & \multicolumn{1}{c|}{} & & \multicolumn{1}{c|}{} & & \\

        - Bias-Loss \cite{huang2025neuralmodelcontextualbiasing} &
        10.0 & \multicolumn{1}{c|}{19.0} &
        98.72 & \multicolumn{1}{c|}{95.10} &
        99.29 & 97.27 \\

        - MPD-Loss (Our) &
        \textbf{9.0} & \multicolumn{1}{c|}{16.9} &
        98.76 & \multicolumn{1}{c|}{95.65} &
        99.43 & 97.55 \\

        - DPD-Loss (Our) &
        \textbf{9.0} & \multicolumn{1}{c|}{\textbf{13.0}} &
        \textbf{99.09} & \multicolumn{1}{c|}{\textbf{96.59}} &
        \textbf{99.60} & \textbf{98.17} \\

        \bottomrule
      \end{tabular}
    }
\end{table}
Table \ref{tab:MainRetrieval} compares the performance of biasing scoring across different objectives.
Unlike the baseline Bias-Loss, which must be implemented as a joint objective (discriminative and log loss) \cite{huang2025neuralmodelcontextualbiasing} to prevent training collapse on multi-target data, our proposed MPD-Loss and DPD-Loss converge effectively as standalone objectives.
Notably, DPD-Loss achieves a Recall\#20 of 99.09\% on test-clean, signifying that target entities are almost invariably ranked within the top-20 candidates.
In contrast, prompt-based baselines like CTC-Filter and K-Prompt suffer from error propagation due to their reliance on initial transcriptions, leading to significantly degraded performance on the challenging test-other.

\begin{table*}[htbp]
    \caption{B-WER (WER/U-WER) performance of COALA and baselines across various biasing list sizes ($N$). Results for the baselines are from their original reports. w/o Biasing represents performance without biasing list information. OOM (out-of-memory) indicates that the required inference VRAM exceeded 24 GB, and BTI is the biasing target identification.}
    \label{tab:MainASR}
    \centering
    \resizebox{\linewidth}{!}{
    \begin{tabular}{lccccccccccc}

      \toprule
      \multicolumn{1}{c}{\multirow{2}{*}{\textbf{Methods}}} &
      \multicolumn{2}{c|}{\textbf{w/o Biasing}} &
      \multicolumn{2}{c|}{\textbf{Oracle}} &
      \multicolumn{2}{c|}{\textbf{$N$=500}} &
      \multicolumn{2}{c|}{\textbf{$N$=1000}} &
      \multicolumn{2}{c}{\textbf{$N$=5000}} \\
      \cmidrule{2-11}
        &
      \textbf{clean} & \multicolumn{1}{c|}{\textbf{other}} &
      \textbf{clean} & \multicolumn{1}{c|}{\textbf{other}} &
      \textbf{clean} & \multicolumn{1}{c|}{\textbf{other}} &
      \textbf{clean} & \multicolumn{1}{c|}{\textbf{other}} &
      \textbf{clean} & \textbf{other} \\
      \midrule

      Bias Qwen \cite{gong2024contextual} &
      \makecell{\textbf{8.40} \\ (2.00 / 1.30)} & \multicolumn{1}{c|}{\makecell{\textbf{18.40} \\ (4.20 / 2.60)}} &
      - & \multicolumn{1}{c|}{-} & 
      \makecell{6.00 \\ (1.90 / 1.40)} & \multicolumn{1}{c|}{\makecell{14.20 \\ (3.90 / 2.70)}} &
      - & \multicolumn{1}{c|}{-} & 
      - & - \\

      RNN-T + IB \cite{shakeel2024contextualized} &
      \makecell{12.96 \\ (2.85 / 1.60)} & \multicolumn{1}{c|}{\makecell{28.09 \\ (7.18 / 4.80)}} &
      - & \multicolumn{1}{c|}{-} & 
      \makecell{11.23 \\ (2.80 / 1.62)} & \multicolumn{1}{c|}{\makecell{26.30 \\ (7.10 / 4.91)}} &
      \makecell{12.44 \\ (2.84 / 1.66)} & \multicolumn{1}{c|}{\makecell{27.93 \\ (7.34 / 5.00)}} &
      - & - \\

      \cite{8068205} + BPB \cite{sudo2024contextualized} &
      \makecell{14.10 \\ (5.05 / 3.90)} & \multicolumn{1}{c|}{\makecell{27.90 \\ (8.81 / 6.60)}} &
      - & \multicolumn{1}{c|}{-} & 
      \makecell{7.00 \\ (3.21 / 2.70)} & \multicolumn{1}{c|}{\makecell{13.50 \\ (6.28 / 5.50)}} &
      \makecell{7.70 \\ (3.47 / 3.00)} & \multicolumn{1}{c|}{\makecell{15.80 \\ (7.34 / 6.40)}} &
      - & - \\

      \midrule

      COALA &
      \makecell{23.39 \\ (4.54 / 2.34)} & \multicolumn{1}{c|}{\makecell{39.49 \\ (8.75 / 5.48)}} &
      \makecell{\textbf{1.77} \\ (1.90 / 1.92)} & \multicolumn{1}{c|}{\makecell{\textbf{6.12} \\ (4.86 / 4.73)}} &
      \makecell{20.38 \\ (4.36 / 2.49)} & \multicolumn{1}{c|}{\makecell{35.01 \\ (8.87 / 6.09)}} &
      \makecell{21.98 \\ (4.56 / 2.53)} & \multicolumn{1}{c|}{\makecell{37.54 \\ (9.18 / 6.16)}} &
      OOM & OOM \\

      + BTI (DPD-Loss) &
      \makecell{23.39 \\ (4.54 / 2.34)} & \multicolumn{1}{c|}{\makecell{39.49 \\ (8.75 / 5.48)}} &
      \makecell{\textbf{1.77} \\ (1.90 / 1.92)} & \multicolumn{1}{c|}{\makecell{\textbf{6.12} \\ (4.86 / 4.73)}} &
      \makecell{\textbf{3.25} \\ (2.28 / 2.17)} & \multicolumn{1}{c|}{\makecell{\textbf{9.13} \\ (5.57 / 5.20)}} &
      \makecell{\textbf{3.86} \\ (2.50 / 2.34)} & \multicolumn{1}{c|}{\makecell{\textbf{10.59} \\ (5.98 / 5.49)}} &
      \makecell{\textbf{6.96} \\ (3.39 / 2.97)} & \makecell{\textbf{15.17} \\ (7.33 / 6.50)} \\

      \bottomrule
    \end{tabular}
    }
\end{table*}

\begin{figure*}[htbp]
    \begin{minipage}[t]{0.3534\linewidth}
        \centering
        \includegraphics[width=\linewidth]{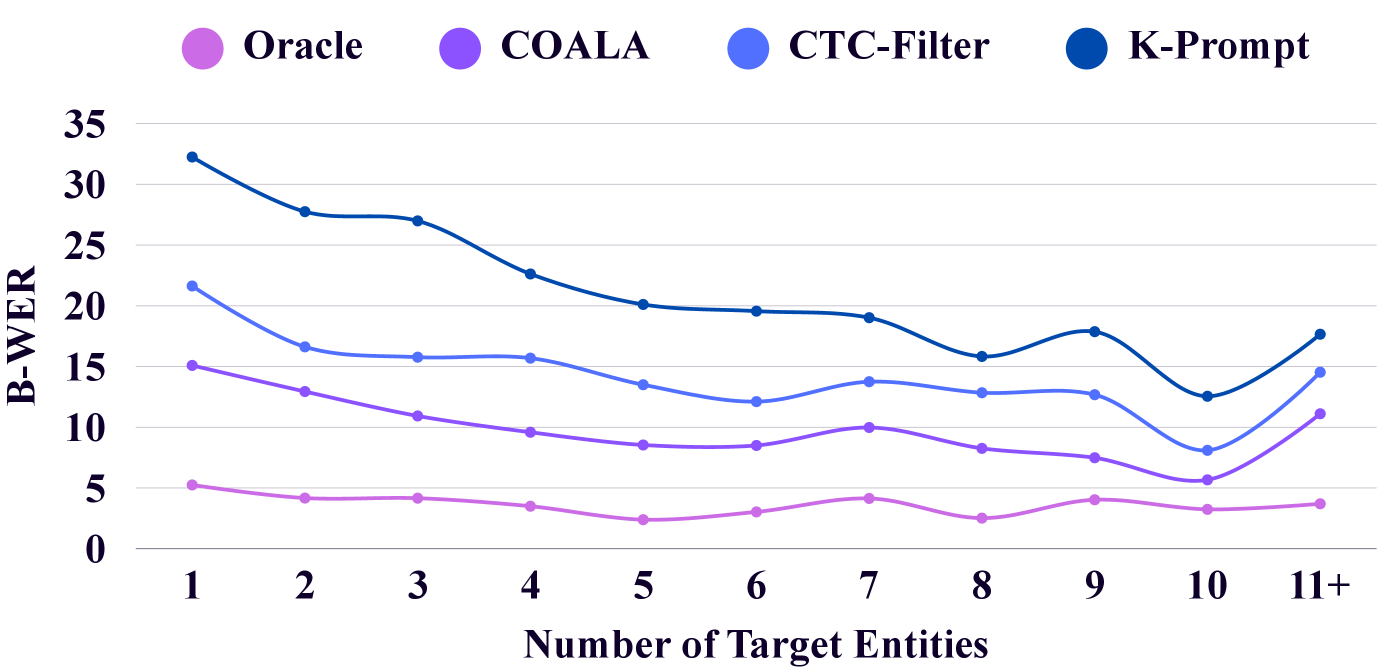}
        \caption{B-WER of COALA-based ASR using biasing prompts from various filter methods on testsets across different numbers of target entities. The COALA is optimized via DPD-Loss.}
        \label{fig:bwer}
    \end{minipage}
    \hfill
    \begin{minipage}[t]{0.6241\linewidth}
        \centering
        \includegraphics[width=\linewidth]{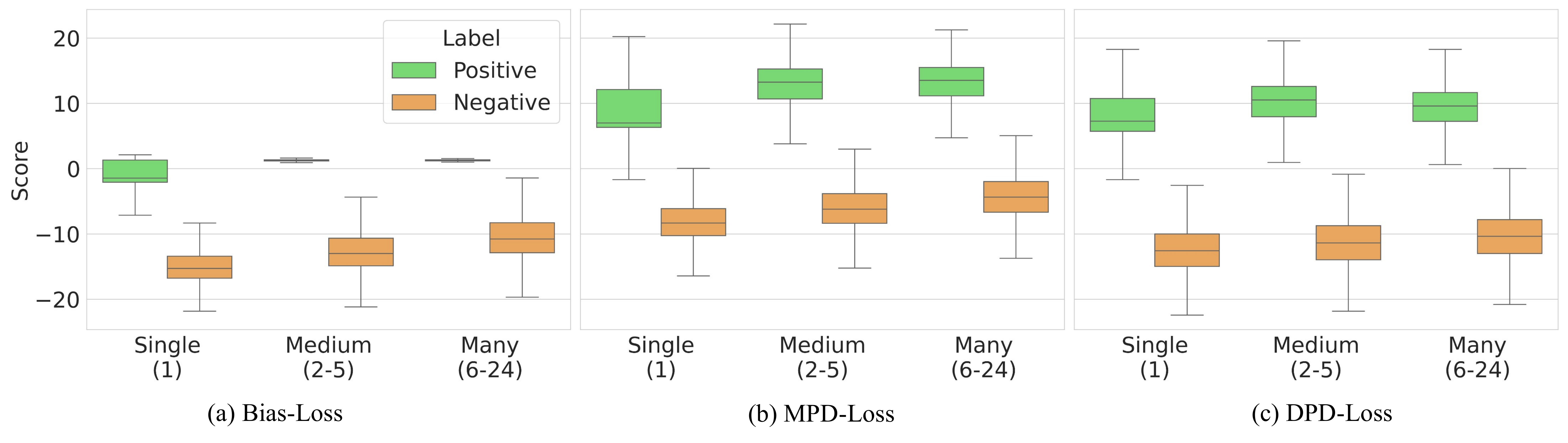}
        \caption{Distribution of biasing scores for positive and negative entities across different objective functions. Scores are categorized by the number of target entities per utterance: single (1), medium (2-5), and many (6-24). The Bias-Loss is a joint objective that combination of the discriminative loss and log loss.}
        \label{fig:boxplot}
    \end{minipage}
\end{figure*}

\subsection{The Performance of Contextualized ASR}

Table \ref{tab:MainASR} presents a comparative evaluation of COALA against several competitive baselines, with reference results cited from their original reports.
While COALA exhibits an inherent performance disparity in the unbiased configuration—specifically trailing by 9.29\% and 11.4\% B-WER on test-clean and test-other—it achieves a decisive performance turnaround under contextual biasing conditions.
Our analysis reveals that without biasing target identification stage and directly feeding the fully biasing list into the SLM prompt is not only marginally effective but also computationally prohibitive.
As the list size $N$ reaches 5000, this unfiltered approach inevitably triggers out-of-memory (OOM) errors due to the limited context-window constraints of the backbone language model.
In contrast, by integrating our biasing target identification mechanism optimized via DPD-Loss, COALA successfully identify target entities from biasing list and improved B-WER performance.
Furthermore, Figure \ref{fig:bwer} illustrates the B-WER fluctuations relative to the number of target entities per utterance.
Notably, since we adopt a top-10 selection strategy, COALA is theoretically incapable of capturing all target entities in utterances containing more than 11 entities.
Despite this inherent constraint, COALA consistently maintains superior performance compared to CTC-Filter and K-Prompt in these high-density scenarios.
This robustness underscores the high precision of COALA in ranking the target entities, ensuring that even a partial retrieval significantly benefits the subsequent ASR task.

\subsection{Analysis of Biasing Scoring}

To further investigate the impact of different objectives on the discriminative capability of the scoring mechanism, we visualize the distribution of biasing scores in Figure \ref{fig:boxplot}.
As shown in Figure \ref{fig:boxplot} (a), the model trained with the baseline Bias-Loss fails to sufficiently elevate positive entity scores.
This deficiency is likely due to the global normalization constraints of discriminative loss, where the auxiliary log loss is not explicitly aligned with the scoring objective.
While MPD-Loss (Figure \ref{fig:boxplot} (b)) successfully prioritizes positive candidates, it struggles to suppress negative entity scores.
We attribute this to its formulation as a relative ranking objective: it ensures positive scores exceed negative ones but lacks the mechanism to calibrate their absolute magnitudes, resulting in an ambiguous decision boundary.
In contrast, DPD-Loss (Figure \ref{fig:boxplot} (c)) demonstrates superior discriminative power by simultaneously maximizing positive scores and minimizing negative ones.
By reframing biasing retrieval as a point-wise binary classification task, DPD-Loss establishes a stable decision boundary centered around zero.

\section{Conclusion}

In this paper, we presented COALA, a robust contextual biasing framework tailored for SLMs.
To address the limitations of traditional discriminative losses in multi-entity scenarios—namely, the gradient conflicts that hinder the optimization of multiple positive targets—we introduced MPD-Loss and DPD-Loss.
Our analysis confirms that by reframing biasing scoring as a point-wise binary classification task, DPD-Loss establishes a stable, absolute decision boundary that enhances the discriminative capability of the scoring mechanism.
Experimental results on the LibriSpeech corpus demonstrate that the proposed biasing target identification mechanism effectively identify target entities from large-scale biasing list.
This precise biasing scoring performance translates into ASR improvements under contextual biasing, significantly reducing B-WER across the testsets.

\section{Acknowledgments}
This work was supported in part by Realtek Semiconductor Corporation under Grant Numbers 113KK01103 and 114KK01005. Any findings and implications in the paper do not necessarily reflect those of the sponsors.

\section{Generative AI Use Disclosure}
The authors utilized Gemini-3.1-Pro to refine the wording clarity of this manuscript. We maintained full control over the research content and remain responsible for the accuracy and integrity of the experimental results and the presented figures.

\bibliographystyle{IEEEtran}
\bibliography{main}

\end{document}